\documentclass[runningheads]{llncs}
\usepackage[T1]{fontenc}
\usepackage{graphicx}
\usepackage{subcaption}
\usepackage{booktabs}
\usepackage{multirow}
\usepackage{array}
\usepackage{xcolor}
\usepackage{makecell}
\usepackage{pifont}
\usepackage{amsmath}
\usepackage{hyperref}
\usepackage{amssymb}

\begin{document}
\title{SRA-Seg: Synthetic to Real Alignment for Semi-Supervised Medical Image Segmentation}
\titlerunning{SRA-Seg}
\author{OFM Riaz Rahman Aranya\inst{1}\orcidID{0000-0002-8195-2710} \and
Kevin Desai\inst{1}\orcidID{0000-0002-2964-8981}}
\authorrunning{O. R. R. Aranya \and K. Desai}
\institute{The University of Texas at San Antonio, TX 78249, USA \\
\email{\{ofmriazrahman.aranya, kevin.desai\}@utsa.edu}}
\maketitle              %
\begin{abstract}
Synthetic data, an appealing alternative to extensive expert-annotated data for medical image segmentation, consistently fails to improve segmentation performance despite its visual realism. The reason being that synthetic and real medical images exist in different semantic feature spaces, creating a domain gap that current semi-supervised learning methods cannot bridge. We propose SRA-Seg, a framework explicitly designed to align synthetic and real feature distributions for medical image segmentation. SRA-Seg introduces a similarity-alignment (SA) loss using frozen DINOv2 embeddings to pull synthetic representations toward their nearest real counterparts in semantic space. We employ soft edge blending to create smooth anatomical transitions and continuous labels, eliminating the hard boundaries from traditional copy-paste augmentation. The framework generates pseudo-labels for synthetic images via an EMA teacher model and applies soft-segmentation losses that respect uncertainty in mixed regions. Our experiments demonstrate strong results: using only 10\% labeled real data and 90\% synthetic unlabeled data, SRA-Seg achieves 89.34\% Dice on ACDC and 84.42\% on FIVES, significantly outperforming existing semi-supervised methods and matching the performance of methods using real unlabeled data. 

Code is available at 
\href{https://github.com/UTSA-VIRLab/SRA-Seg}{https://github.com/UTSA-VIRLab/SRA-Seg}

\end{abstract}

\vspace{-10mm}
\section{Introduction}
\vspace{-3mm}
Using deep learning for medical image segmentation typically requires large, carefully annotated datasets to achieve state-of-the-art performance \cite{Litjens_2017}. Unlike general computer vision, acquiring extensive ground truth in medical imaging is inherently complex, time-consuming, and resource-intensive, requiring specialized professional expertise. This scarcity directly impacts model accuracy, as performance typically scales with dataset size.

To address this data bottleneck, synthetic data generation has emerged as a promising data augmentation strategy \cite{Ibrahim_2025}. However, a significant challenge persists, which is the inherent domain gap between synthetic and real medical images. Directly mixing synthetic and real images during training can cause models to fail on new test data because synthetic images, despite visual plausibility, can introduce artifacts, hallucinations, or lack the intricate, clinically relevant anatomical details crucial for robust segmentation in real-world scenarios \cite{cohen2018distributionmatchinglosseshallucinate}. 

\begin{figure}
  \centering
  \begin{subfigure}[t]{0.22\columnwidth}
    \includegraphics[width=1.0\linewidth]{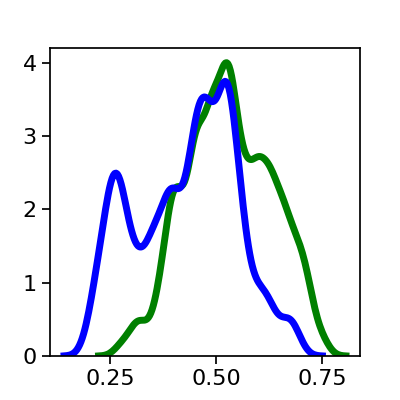}
    \caption{} %
    \label{fig:graph-base-r}
  \end{subfigure}\hspace{3mm}
  \begin{subfigure}[t]{0.22\columnwidth}
    \includegraphics[width=1.0\linewidth]{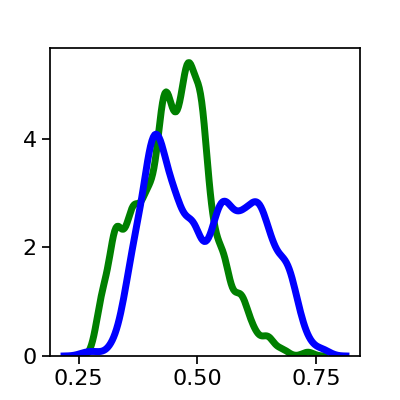}
    \caption{} %
    \label{fig:graph-base-s}
  \end{subfigure}\hspace{3mm}
  \begin{subfigure}[t]{0.22\columnwidth}
    \includegraphics[width=1.0\linewidth]{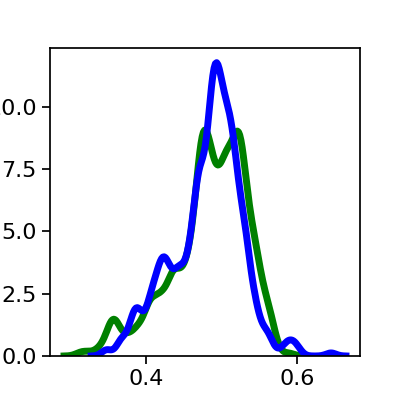}
    \caption{} %
    \label{fig:graph-bcp}
  \end{subfigure}\hspace{3mm}
  \begin{subfigure}[t]{0.22\columnwidth}
    \includegraphics[width=1.0\linewidth]{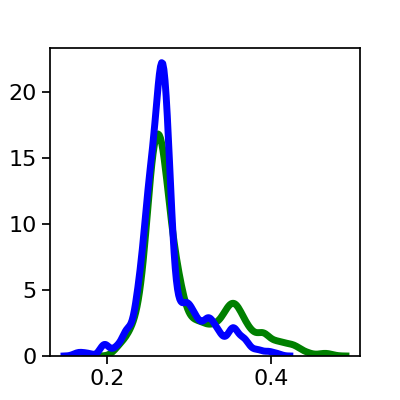}
    \caption{} %
    \label{fig:graph-SA-BCP}
  \end{subfigure}
  \vspace{-3mm}
  \caption{Kernel Density Estimation (KDE) graphs showcasing domain mismatch between labeled (green line) and unlabeled (blue line) data for ``Right Ventricle'' class of the ACDC dataset \cite{acdc}. \textbf{(a)} UNet\cite{unet} with Real Unlabeled Data \textbf{(b)} UNet\cite{unet} with Synthetic Unlabeled Data \textbf{(c)} BCP\cite{bai2023bidirectionalcopypastesemisupervisedmedical} with Synthetic Unlabeled Data \textbf{(d)} Proposed SRA-Seg with Synthetic Unlabeled Data.} 
  \vspace{-6mm}
  \label{fig:graphs}
\end{figure}

This domain gap problem exists in both synthetic and real-world datasets, especially when data is aggregated from various sources. It causes significant distributional disparities between labeled and unlabeled samples, making model training more challenging. As shown in Figure \ref{fig:graph-base-r}, a kernel density estimation plot of a baseline UNet \cite{unet} on the Right Ventricle class of the ACDC dataset \cite{acdc} clearly reveals this inherent domain shift.

The challenge intensifies when synthetic data is integrated into semi-supervised learning (SSL) frameworks. Studies like SinGAN-Seg \cite{singan} and recent work with diffusion models \cite{saragih2024usingdiffusionmodelsgenerate,Ji_Diffusionbased_MICCAI2024} consistently highlight that while synthetic images may appear visually convincing, they often lack the nuanced, intricate features of genuine medical images, leading to reduced generalization capabilities \cite{Ji_Diffusionbased_MICCAI2024,Hu2021}. For instance, Figure \ref{fig:graph-base-s} demonstrates an increased domain gap between real labeled data and corresponding synthetic unlabeled data generated for the ACDC dataset using StyleGAN2-ADA \cite{karras2020traininggenerativeadversarialnetworks}.

Generative Adversarial Networks (GANs) \cite{goodfellow2014generativeadversarialnetworks} have revolutionized synthetic data generation, but their practical utility for medical image segmentation remains limited by a persistent domain gap \cite{kamnitsas2016unsuperviseddomainadaptationbrain,li2023medicalimagesegmentationdomain}.
CycleGAN \cite{CycleGAN2017} excels at domain-to-domain translation, while SinGAN-Seg \cite{singan} addresses data scarcity by generating multiple samples from a single input.
StyleGAN2-ADA \cite{karras2020traininggenerativeadversarialnetworks} addresses overfitting on small medical datasets by dynamically adjusting augmentations, unlike StyleGAN \cite{stylegan} which struggled with it. StyleGAN2-ADA's ability to produce high-quality synthetic images with fewer real training samples makes it particularly well-suited for medical image generation. However, the synthetic-real domain gap remains a significant challenge in SSL, exacerbated by substantial disparities between labeled and unlabeled data, especially when the unlabeled data includes synthetic samples. This disparity leads to substantial empirical mismatches\cite{li2025semi,li2023medicalimagesegmentationdomain}. While the Bidirectional Copy-Paste (BCP) method \cite{bai2023bidirectionalcopypastesemisupervisedmedical} successfully reduces the distributional gap between real labeled and real unlabeled datasets, it is not optimized for scenarios involving synthetic data, as illustrated by its inefficacy in bridging this gap in Figure \ref{fig:graph-bcp}.

\textbf{Proposed Approach:} We propose SRA-Seg that effectively integrates high-quality synthetic images as unlabeled data within semi-supervised learning. Our method introduces a Similarity-Alignment (SA) loss to explicitly align synthetic feature distributions with their closest real counterparts, effectively bridging the synthetic-to-real domain gap, as shown in Figure \ref{fig:graph-SA-BCP}. Through this principled alignment strategy, SRA-Seg achieves superior segmentation performance, demonstrating that synthetic data can match or exceed the utility of real unlabeled data when properly aligned.

In summary, our contributions are:
\vspace{-2mm}
\begin{itemize}
\item A framework explicitly designed to bridge the synthetic-to-real domain gap for semi-supervised medical image segmentation.
\item A similarity-alignment (SA) loss using frozen DINOv2 embeddings to pull synthetic features toward real counterparts.
\item Soft edge blending for smooth anatomical transitions, replacing hard copy-paste boundaries.
\item EMA-based one-hot pseudo-label generation enabling effective use of synthetic images in SSL.
\end{itemize}

\vspace{-5mm}
\section{Related Work}
\vspace{-3mm}
Many conventional SSL methods, like self-training and Mean Teacher frameworks \cite{mean-teacher}, mostly use pseudo-labeling and consistency losses. While methods like SASSNet \cite{Li_2020}, UA-MT \cite{yu2019uncertaintyawareselfensemblingmodelsemisupervised}, and UMCT \cite{umct} incorporate geometric constraints and uncertainty awareness to better leverage unlabeled data, they often treat labeled and unlabeled samples independently, missing valuable inter-domain information and struggling to effectively integrate synthetic data. Similarly, data augmentation strategies like CutMix \cite{Yun2019CutMix} enhance data diversity, but still face performance limitations when pseudo-labels provide imprecise supervision.

Recent advances in semi-supervised medical segmentation show improved performance but still lack synthetic data handling capabilities. CrossMatch \cite{crossmatch} combines perturbation strategies with knowledge distillation. DiffRect \cite{diffrect} leverages latent diffusion models for pseudo-label refinement. ADB \cite{abd} introduces adaptive displacement fields that adjust based on local image characteristics, and CGS \cite{cgs} proposes a collaborative framework for multi-target segmentation. BCP \cite{bai2023bidirectionalcopypastesemisupervisedmedical} creates mixed samples through copy-paste operations, enforcing bidirectional consistency. While these methods excel with real data, none provide mechanisms to handle synthetic data or adapt to the distribution gaps between real and generated samples.

Domain adaptation methods have been extensively studied to address distribution shifts in medical imaging \cite{li2023medicalimagesegmentationdomain}. Classical approaches like adversarial domain adaptation \cite{kamnitsas2016unsuperviseddomainadaptationbrain} and recent methods including dual-level alignment \cite{li2025semi} and diffusion-based adaptation \cite{Ji_Diffusionbased_MICCAI2024} show promise but typically require domain labels and struggle with continuous quality variations present in synthetic data. 
Disentanglement learning approaches \cite{higgins2017betavae,huang2018munit,chartsias2019sdnet} attempt to separate domain-invariant content from domain-specific style. Recent theoretical work \cite{locatello2019challenging} demonstrated that unsupervised disentanglement is fundamentally impossible without strong inductive biases, requiring extensive hyperparameter tuning that doesn't generalize across datasets. Furthermore, medical imaging studies \cite{MULLER2025103628} confirm that pathological changes simultaneously affect appearance and structure, violating the separable content and style assumption. Synthetic data generation artifacts contain unrealistic feature combinations that don’t follow the content-style dichotomy. These methods also require adversarial training or VAE frameworks with multiple encoders/decoders, increasing computational cost and training instability compared to our streamlined approach. In contrast, SRA-Seg’s direct feature alignment avoids these limitations while remaining simple and effective.

Current SSL methods exhibit critical limitations when integrating synthetic data: 
(1) inability to handle generative model outputs, (2) uniform treatment regardless 
of data quality, (3) lack of adaptive mechanisms for distribution gaps, and 
(4) insufficient focus on boundary regions. SRA-Seg addresses these gaps through 
adaptive feature alignment.

\vspace{-3mm}
\section{Methodology}
\vspace{-3mm}

\begin{figure}[t]  %
    \centering
    \includegraphics[width=\linewidth]{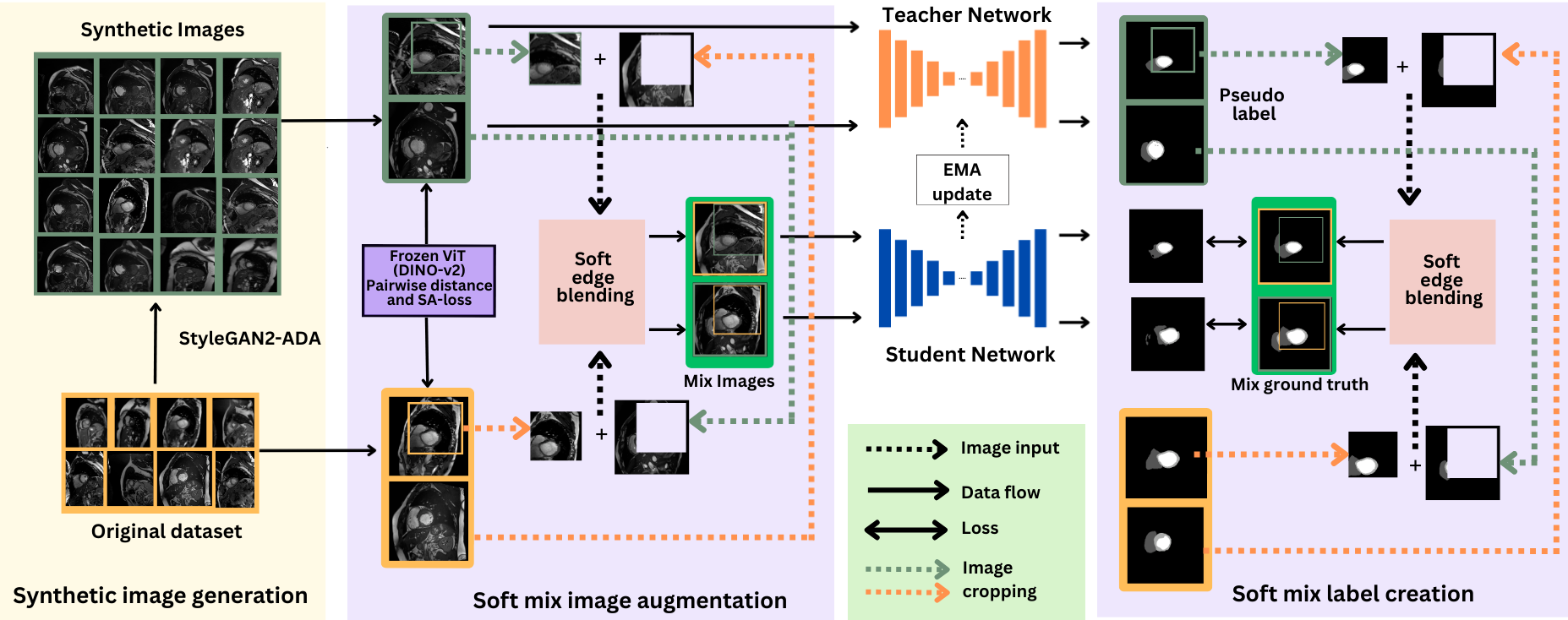}  %
    \vspace{-5mm}
    \caption{Overview of the proposed SRA-Seg method}
    \label{fig:SA-BCP}  %
\vspace{-5mm}
\end{figure}

The proposed SRA-Seg method addresses data scarcity in medical image segmentation by integrating synthetic unlabeled images into SSL. Figure \ref{fig:SA-BCP} outlines the SRA-Seg's workflow, with subsections detailing each component.

\vspace{-6mm}
\subsection{Dataset Preparation}
\vspace{-2mm}
\subsubsection{Real Labeled Data}
Our methodology begins with a limited subset of the original real dataset, designated as the real labeled images \(V_{\mathrm{lab}}\). To evaluate SRA-Seg under data-scarce conditions, we restrict this labeled data to only 5\% or 10\% of the total available samples. These specific proportions are chosen to align with the experimental setups of existing semi-supervised methods, ensuring a fair and consistent basis for comparative performance analysis.

\vspace{-3mm}

\vspace{-2mm}
\subsubsection{Synthetic Unlabeled Data Generation} \label{sec:syn-data-gen}
\begin{figure}[t]  %
    \centering
    \includegraphics[width=\linewidth]{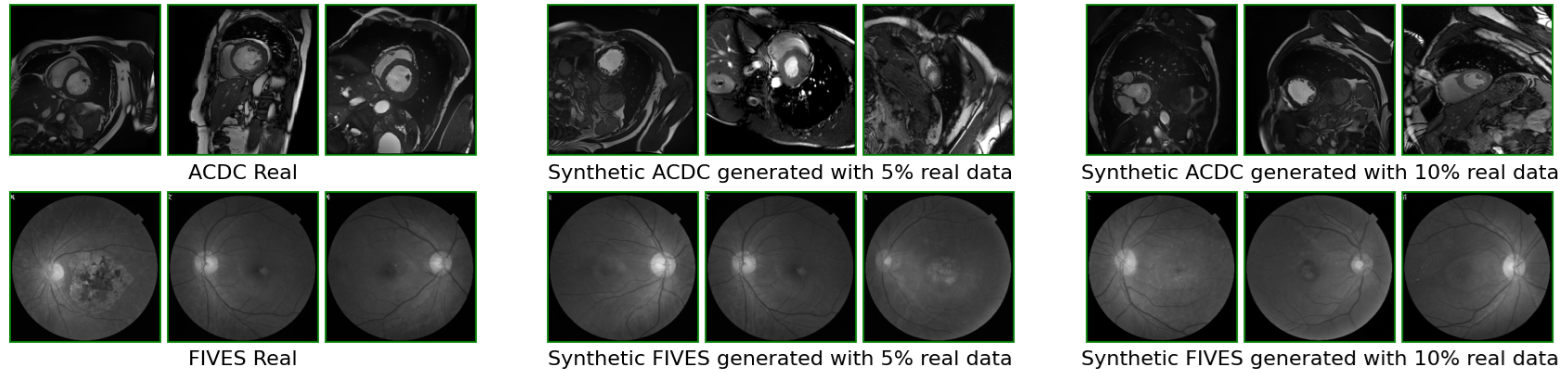}  %
    \vspace{-5mm}
    \caption{Synthetic Data Generated by the StyleGAN2-ADA\cite{karras2020traininggenerativeadversarialnetworks} for the ACDC dataset\cite{acdc} (top) and the FIVES dataset\cite{Jin2022} (bottom). Left to right - the first 3 are the original images, the next 3 are the images generated using 5\% real data during training, and the last 3 images are generated using 10\% real data.}
    \vspace{-5mm}
\label{fig:synthetic_data}  %
\end{figure}

To augment the limited real labeled data, high-fidelity synthetic images \(V_{\mathrm{syn}}\) are generated using StyleGAN2-ADA \cite{karras2020traininggenerativeadversarialnetworks}. The model's Adaptive Discriminator Augmentation (ADA) mechanism dynamically adjusts augmentation probability to prevent discriminator overfitting on limited training samples, enabling realistic and diverse image generation even from minimal input. Figure \ref{fig:synthetic_data} presents random selected synthetic images for the ACDC and FIVES datasets at the 5\% and 10\% labeled data splits.

We train StyleGAN2-ADA on the aforementioned 5\% or 10\% real labeled data. The quantity of generated synthetic images matches the remaining proportion of the full dataset (e.g., 90\% or 95\%), effectively serving as the unlabeled data input for the SRA-Seg framework.

\vspace{-4mm}
\subsection{SRA-Seg Framework}
\vspace{-1mm}
\subsubsection{Synthetic Pseudo-Label Generation}
Pseudo-labels for each synthetic (unlabeled) image \(V_{\mathrm{syn}}\) are produced by a teacher network maintained as an exponential moving average (EMA) of the student network’s weights. After each update of the student parameters \(\theta_{\mathrm{student}}\), the teacher parameters \(\theta_{\mathrm{teacher}}\) are updated via
\vspace{-1mm}
\small
\begin{equation}
  \theta_{\mathrm{teacher}}
  \;\leftarrow\;
  \alpha\,\theta_{\mathrm{teacher}}
  \;+\;
  (1-\alpha)\,\theta_{\mathrm{student}},
  \quad
  \alpha\in[0,1)
\end{equation}
\vspace{-1mm}
\normalsize
The EMA teacher takes each synthetic slice \(V_{\mathrm{syn}}\) and produces raw, per‐class logits \(\ell_c(x,y)\). These logits are then normalized via the softmax function, resulting in the per‐class probability maps \(p_c(x,y)\)
\vspace{-1mm}
\small
\begin{equation}
  p_c(x,y)
  = 
  \frac{\exp\!\bigl(\ell_c(x,y)\bigr)}
       {\sum_{k=1}^C \exp\!\bigl(\ell_k(x,y)\bigr)},
  \quad
  c=1,\dots,C
\end{equation}
\vspace{-1mm}
\normalsize

\noindent Discrete pseudo-labels \(\hat{\ell}(x,y)\) are obtained by assigning each pixel to its most likely class. 
\vspace{-1mm}
\small
\begin{equation}
  \hat{\ell}(x,y)
  = 
  \arg\max_{c=1,\dots,C}\;p_c(x,y)
\end{equation}
\vspace{-1mm}
\normalsize
Spurious predictions are removed by applying a two-dimensional largest-connected-component filter to each class mask 
\(\mathbf m_c(x,y)=\mathbf1\{\hat{\ell}(x,y)=c\}\), retaining only the largest component per class. The resulting cleaned map \(\hat{\ell}(x,y)\) is converted to a one-hot tensor
\vspace{-2mm}
\small
\begin{equation}
  L_{\mathrm{pseudo}}(c,x,y)
  =
  \begin{cases}
    1, & \hat{\ell}(x,y)=c\\
    0, & \text{otherwise}
  \end{cases}
\end{equation}
\vspace{-1mm}
\normalsize
serves as the pseudo-label for subsequent augmentation and loss computations.

\vspace{-2mm}
\subsubsection{Soft-Mix Augmentation} \label{sec:soft-mix}
To reduce the domain gap between real and synthetic data, we introduce soft-mix augmentation with bidirectional patch exchange. Current methods like BCP~\cite{bai2023bidirectionalcopypastesemisupervisedmedical} and ADB~\cite{abd} use direct copy-paste operations that create sharp boundaries, degrading segmentation performance. Our soft blending strategy produces smooth anatomical transitions for learning generalizable features. We generate a blending mask by randomly selecting a rectangular region spanning two-thirds of the image dimensions, with the hole region initialized to 0 and the outer region to 1.
\vspace{-1mm}
\small
\begin{equation}
  m(x,y)=
  \begin{cases}
    0,& (x,y)\in\text{hole}\\
    1,& \text{otherwise}
  \end{cases}
\end{equation}
\vspace{-1mm}

\begin{figure}[t]
    \centering
    \includegraphics[width=\linewidth]{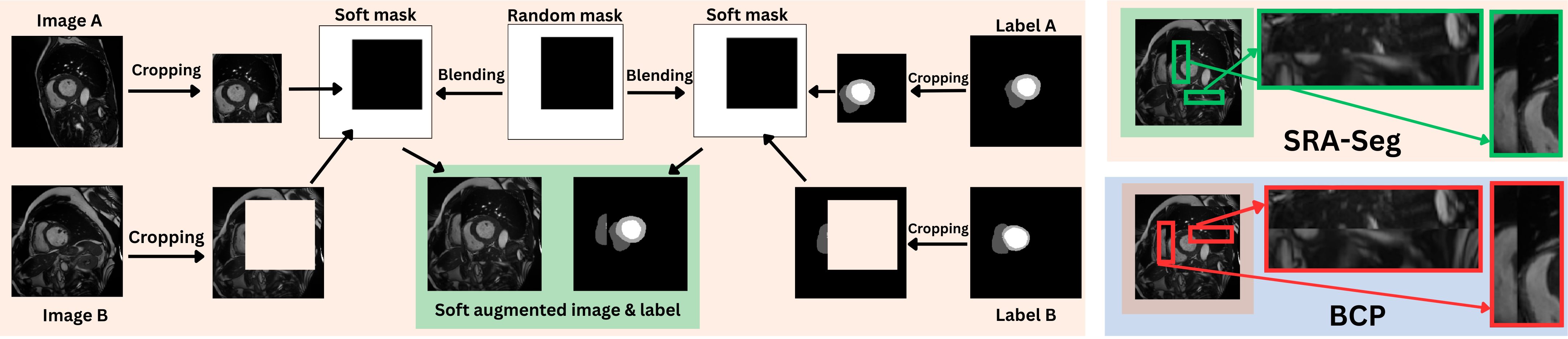}  %
    \vspace{-5mm}
    \caption{Soft Edge Blending component of SRA-Seg. The labeled and unlabeled images as well as the corresponding segmentation masks are mixed through a cropping and soft mask blending to reduce the sharp edges. A zoomed-in comparison of blending results between BCP\cite{bai2023bidirectionalcopypastesemisupervisedmedical} and SRA-Seg is shown.}
    \vspace{-5mm}
    \label{fig:SA-BCP-comp}  %
\end{figure}

\normalsize
\noindent To blur the edges and create a smooth mask, we use a $3\times3$ average-pool (other blending methods are reported in the supplementary material). Portions of the labeled image $V_{\mathrm{lab}}(x,y)$ and the pseudo-labeled synthetic image $V_{\mathrm{syn}}(x,y)$ are cropped at the same spatial location, then blended by $\alpha$:

\vspace{-3mm}
\small
\begin{align}
\widetilde V_1(x,y)
  &= \alpha\,\odot\,V_{\rm syn}(x,y)
    + \bigl[1-\alpha\bigr]\,\odot\,V_{\rm lab}(x,y)\\
\widetilde V_2(x,y)
  &= \alpha\,\odot\,V_{\rm lab}(x,y)
    + \bigl[1-\alpha\bigr]\,\odot\,V_{\rm syn}(x,y)\\
\widetilde L_1(c,x,y)
  &= \alpha\,\odot\,L_{\rm pseudo}(c,x,y)
    + \bigl[1-\alpha\bigr]\,\odot\,L_{\rm lab}(c,x,y)\\
\widetilde L_2(c,x,y)
  &= \alpha\,\odot\,L_{\rm lab}(c,x,y) + \bigl[1-\alpha\bigr]\,\odot\,L_{\rm pseudo}(c,x,y)
\end{align}
\vspace{1mm}
\normalsize

\vspace{-4mm}
\noindent For instance, if $V_{\mathrm{syn}}(x,y)=100$, $V_{\mathrm{lab}}(x,y)=50$, and $\alpha(x,y)=0.6$, then
\vspace{-1mm}
\small
\[
  \widetilde V(x,y) = 0.6\times100 + 0.4\times50 = 80
\vspace{-1mm}
\]
\normalsize
Two complementary mixtures are generated per sample to further diversify the training set.
\vspace{-1mm}
\small
\begin{align}
  \widetilde V_1
    &= \alpha\,\odot\,V_{\mathrm{syn}}
      + \bigl[1-\alpha\bigr]\,\odot\,V_{\mathrm{lab}}\\
  \widetilde V_2
    &= \alpha\,\odot\,V_{\mathrm{lab}}
      + \bigl[1-\alpha\bigr]\,\odot\,V_{\mathrm{syn}}
\end{align}
\vspace{-3mm}
\normalsize

\noindent with corresponding soft labels $\widetilde L_1,\widetilde L_2$. These softly blended pairs serve as inputs and targets for the subsequent loss functions, producing smooth boundary transitions essential for accurate segmentation at anatomical boundaries.

\vspace{-3mm}
\subsection{Loss Functions}
\vspace{-3mm}
To ensure accurate pixel-wise segmentation on soft-mixed images and labels, we introduce two complementary losses: (1) Soft-Segmentation Loss, which promotes precise boundary localization by computing Dice overlap and cross-entropy directly on continuous probability maps, respecting the model’s uncertainty in mixed regions; and (2) Similarity-Alignment (SA) Loss, which shrinks the domain gap by aligning the feature distributions of mixed/unlabeled patches with those of real/labeled patches.  Details are given below.

\vspace{-4mm}
\subsubsection{Soft‐Segmentation Loss:}
Our targets are continuous per-pixel probability maps (not hard one-hots), so we replace discrete Dice and cross-entropy loss with “soft” variants that compute overlap and log-loss directly on these distributions. The Soft‐Segmentation Loss balances precise boundary alignment with well‐calibrated confidence estimates on soft‐mixed examples. It is defined as
\small
\begin{equation}
\mathcal L_{\mathrm{soft}}
= \mathcal L_{\mathrm{Dice}} + \mathcal L_{\mathrm{CE}}
\end{equation}
\noindent Here, the \textit{\textbf{Soft‐Dice Loss}} measures the normalized overlap between predicted and target distributions. It is defined by the following equation, where \(P(c,x,y)\) is the predicted probability for class \(c\) at pixel \((x,y)\), and \(\widetilde L(c,x,y)\) is the soft target.
\small
\begin{equation}
\mathcal L_{\mathrm{Dice}} = 1 - \frac{2\,\sum_{c,x,y} P(c,x,y)\,\widetilde L(c,x,y)}
                 {\sum_{c,x,y}P(c,x,y)\;+\;\sum_{c,x,y}\widetilde L(c,x,y)\;+\;\epsilon}
\end{equation}
\vspace{-1mm}

\noindent Whereas, the \textit{\textbf{Soft-Cross-Entropy Loss}} minimizes the Kullback–Leibler divergence between the prediction and the target distribution and is defined as
\vspace{-1mm}
\small
\begin{equation}
\mathcal L_{\mathrm{CE}}
= -\sum_{c,x,y}\widetilde L(c,x,y)\,\log P(c,x,y)
\end{equation}

\vspace{-4mm}
\subsubsection{Similarity-Alignment Loss}
To explicitly close the gap between synthetic and real images, we introduce a Synthetic–Real Alignment (SA) loss. SA-loss operates on deep feature embeddings extracted by a frozen self-supervised vision transformer (ViT), ensuring that every synthetic example is pulled toward its nearest real-image counterpart in feature space.

\vspace{1mm}
Let
\(\{x_{\mathrm{real}}^{(j)}\}_{j=1}^{N},\; x_{\mathrm{real}}^{(j)} \in \mathbb{R}^{C \times H \times W}\) 
\vspace{1mm}
 be a batch of \(N\) real images, and
\(\{x_{\mathrm{syn}}^{(i)}\}_{i=1}^{M}, \quad x_{\mathrm{syn}}^{(i)} \in \mathbb{R}^{C \times H \times W}\) be a batch of \(M\) synthetic images. We pass both sets through a pretrained, frozen ViT-B/16 DINO-v2\cite{dinov2} ViT feature extractor \(\phi(\cdot)\) to obtain \(D\)-dimensional embeddings:
  \vspace{-1mm}
\small
\[
f_{\mathrm{real}}^{(j)} = \phi\bigl(x_{\mathrm{real}}^{(j)}\bigr), 
\quad
f_{\mathrm{syn}}^{(i)} = \phi\bigl(x_{\mathrm{syn}}^{(i)}\bigr)
  \]

\normalsize
DINO-v2, chosen for its off-the-shelf, semantically rich 2D patch embeddings, directly transfers to medical segmentation without in-domain fine-tuning, unlike models like TransUNet \cite{transunet} or Swin UNETR \cite{swinunet} that require extensive volume-based adaptation. DINO-v2’s self-supervised ViT-B/16 features, pretrained on 142 M images, are proven robust across visual domains, focusing alignment on true anatomy rather than superficial artifacts \cite{dinov2}.

For each synthetic embedding \(f_{\mathrm{syn}}^{(i)}\), we compute the Euclidean distance to every real embedding and determine the minimum distance, which identifies its closest exemplar.
  \vspace{-2mm}
\small
\begin{equation}
d_i \;=\; \min_{1 \le j \le N}\;\bigl\lVert f_{\mathrm{syn}}^{(i)} - f_{\mathrm{real}}^{(j)}\bigr\rVert_2
\end{equation}
\vspace{-1mm}
\normalsize

\noindent The \textbf{\textit{SA-loss}} is the mean of these minimum distances, minimizing which pulls synthetic features closer to real ones, narrowing the synthetic–real divide.
  \vspace{-1mm}
\small
\begin{equation}
\mathcal{L}_{\mathrm{SA}}
\;=\;
\frac{1}{M}\sum_{i=1}^{M} d_i
\;=\;
\frac{1}{M}
\sum_{i=1}^{M}
\min_{1 \le j \le N}
\bigl\lVert f_{\mathrm{syn}}^{(i)} - f_{\mathrm{real}}^{(j)}\bigr\rVert_2
\end{equation}

\vspace{-2mm}
\subsection{Training Optimization}
\vspace{-2mm}
During each training iteration, the SRA-Seg framework optimizes a comprehensive loss function combining the Soft-Segmentation Loss \(\mathcal L_{\mathrm{soft}}\) and the Similarity-Alignment (SA) Loss \(\mathcal L_{\mathrm{SA}}\).

The network first processes a batch of softly blended image-label pairs, generated as described in Section \ref{sec:soft-mix}, to compute \(\mathcal L_{\mathrm{soft}}\). Concurrently, the same mixed (unlabeled) patches and their corresponding real counterparts within the batch are passed through the frozen ViT-B/16 DINO-v2 feature extractor to obtain embeddings for \(\mathcal L_{\mathrm{SA}}\). These two loss terms are combined into a single scalar:
\vspace{-2mm}
\small
\[
  \mathcal L = \mathcal L_{\mathrm{soft}} + \lambda\,\mathcal L_{\mathrm{SA}}
  \vspace{-2mm}
\]
\normalsize
where \(\lambda\) is a tunable hyperparameter that balances the contribution of segmentation accuracy with feature-space alignment. Gradients from both loss components are backpropagated through the segmentation network (but critically, not through the frozen feature extractor). Model parameters are updated using Stochastic Gradient Descent (SGD) with momentum. This joint optimization strategy ensures that the model not only benefits from the soft-mixed supervision but also learns robust representations that effectively bridge the domain gap between real and synthetic data distributions.

\vspace{-3mm}
\section{Experimental Setup}
\vspace{-3mm}
This section details the experimental design, datasets, evaluation metrics, and the strategy for synthetic data generation employed in evaluating the proposed SRA-Seg method. We utilize a standard UNet~\cite{unet} as our baseline segmentation architecture throughout all experiments. 

\noindent \textbf{Hyperparameters:} Based on extensive experimentation (details in supplementary), we set patch size $\beta = 2/3$, kernel size $\alpha = 3$, and SA weight $\lambda = 0.1$.

\noindent \textbf{Evaluation Metrics:}
To thoroughly evaluate SRA-Seg against state-of-the-art semi-supervised segmentation methods, we employ four standard metrics, consistent with the evaluation protocol of other state-of-the-art models: Dice Score (\%), Jaccard Score (\%), 95\% Hausdorff Distance (95HD, voxel), and Average Surface Distance (ASD, voxel). Higher Dice and Jaccard values indicate superior overlap between predicted and ground-truth masks, while lower ASD and 95HD values signify a tighter and more accurate boundary alignment.

\vspace{-4mm}
\subsection{Datasets}
\vspace{-2mm}
We evaluate our proposed method on two datasets:

\noindent \textbf{ACDC dataset} \cite{acdc} is a four-class segmentation benchmark (background, right ventricle, left ventricle, myocardium) comprising 100 patient scans. Following the setup in BCP and related works, we allocate 70 scans for training (1,312 images), 10 for validation, and 20 for testing. For semi-supervised scenarios, we explore 5\% and 10\% splits of the training data, equivalent to 68 and 136 images respectively, to align with prior semi-supervised methodologies and simulate low-data conditions.

\noindent \textbf{FIVES dataset} \cite{Jin2022} consists of 800 high-quality multi-disease fundus images and annotated segmentation masks. We partition this dataset into 70\% for training (560 images), 10\% for validation, and 20\% for testing. Similar to ACDC, we use 5\% (28 images) and 10\% (56 images) of the 560 training images as labeled data to simulate low-data availability, treating the remaining data as unlabeled.

\noindent \textbf{Synthetic Data:}
For our experiments, high-fidelity synthetic images are generated using StyleGAN2-ADA~\cite{karras2020traininggenerativeadversarialnetworks} as described in Section \ref{sec:syn-data-gen}. The model is trained exclusively on the limited percentage of real labeled data available for each dataset (e.g., 5\% or 10\%). For instance, when using 10\% labeled data (136 images for ACDC or 56 for FIVES), StyleGAN2-ADA is trained solely on these specific images. 
Figure~\ref{fig:synthetic_data} shows random synthetic samples generated for both ACDC and FIVES datasets at 5\% and 10\% labeled data splits.

\vspace{-3mm}
\subsection{Methods Compared}
\vspace{-2mm}
We evaluate SRA-Seg against different baselines depending on the data type. For \textit{real unlabeled data}, we compare against UNet \cite{unet} as a supervised baseline and BCP \cite{bai2023bidirectionalcopypastesemisupervisedmedical} as a recent SSL method. This comparison validates our method's effectiveness even when synthetic data challenges are absent. For \textit{synthetic unlabeled data}, we compare against five recent state-of-the-art SSL methods: BCP \cite{bai2023bidirectionalcopypastesemisupervisedmedical}, CrossMatch \cite{crossmatch}, ABD(BCP) \cite{abd}, DiffRect \cite{diffrect}, and CGS \cite{cgs}.

\vspace{-3mm}
\section{Results}
\vspace{-3mm}
This section presents the comprehensive outcomes of our experiments, demonstrating the performance of SRA-Seg in various data-constrained scenarios. As visually demonstrated in Figure~\ref{fig:graphs}, while incorporating synthetic data can initially widen the domain gap between labeled and unlabeled distributions, SRA-Seg effectively mitigates this gap, thereby enabling medical segmentation models to leverage the full potential of advanced generative algorithms.

\begin{figure}[b]  %
    \centering
    \vspace{-5mm}
    \includegraphics[width=\linewidth]{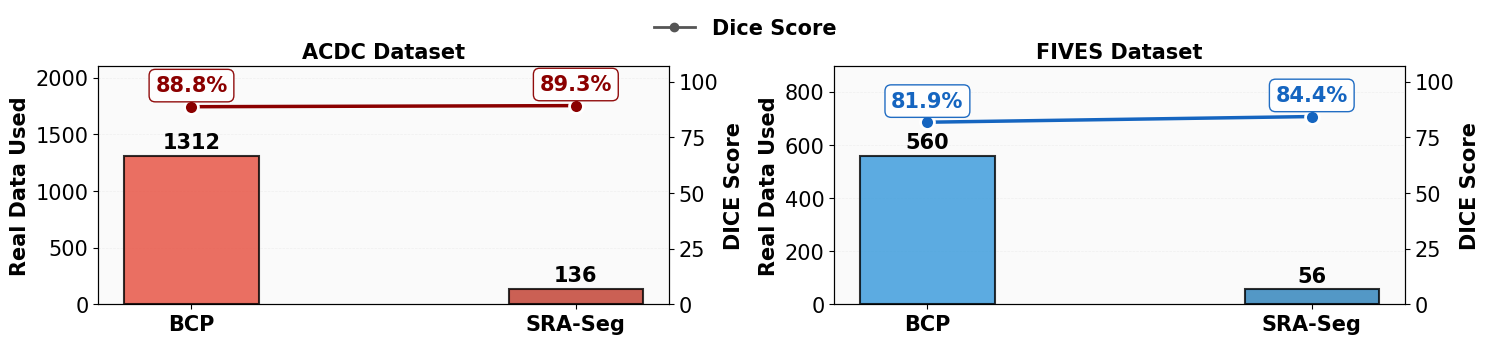}  %
    \vspace{-6mm}
    \caption{Comparison of real‐image usage (bar height) and resulting Dice scores (red markers) for BCP\cite{bai2023bidirectionalcopypastesemisupervisedmedical} versus SRA-Seg on the ACDC\cite{acdc} and FIVES\cite{Jin2022} datasets.}
    \label{fig:dice-comp}  %
\end{figure}

Figure~\ref{fig:dice-comp} gives an overview of SRA-Seg’s performance. It demonstrates that, even when trained with only 10\% real data (and 90\% synthetic), SRA-Seg outperforms the leading model BCP\cite{bai2023bidirectionalcopypastesemisupervisedmedical} by achieving higher Dice scores on both the ACDC\cite{acdc} and FIVES\cite{Jin2022} datasets. Detailed quantitative and qualitative results are presented in following sections.

\vspace{-3mm}
\subsection{Results on ACDC Dataset}
\vspace{-2mm}
\textbf{Quantitative Results:} Table~\ref{table:quant-acdc} presents segmentation results on ACDC dataset~\cite{acdc} across 5\% and 10\% labeled data conditions.
As the results demonstrate, SRA-Seg achieves superior performance compared to all the other models in the 10\% labeled data (with 90\% synthetic unlabeled data) scenario, coming a close second in the 5\% scenario as well. SRA-Seg also obtains better results than the real unlabeled data scenario, demonstrating the potential to significantly reduce the reliance on extensive real data collection.

\begin{table}[t]
\caption{Comparison of our SRA-Seg approach against other semi-supervised segmentation models on the ACDC dataset \cite{acdc}. We mark the best (bold) and second best (underline) results for the synthetic data usage in both 5\% and 10\% data split categories.}
\label{table:quant-acdc}
\centering
\scriptsize %
\setlength{\tabcolsep}{1pt} %
\renewcommand{\arraystretch}{0.8}
\begin{tabular}{l|c|cc|c@{\hspace{6pt}}c@{\hspace{6pt}}c@{\hspace{6pt}}c}
\toprule
\multirow{3}{*}{\textbf{Method}} & \textbf{Labeled} & \multicolumn{2}{c|}{\textbf{Unlabeled}} & \multirow{3}{*}{\makecell{\textbf{DICE}\textbf{↑}}} & \multirow{3}{*}{\makecell{\textbf{Jaccard}\textbf{↑}}} & \multirow{3}{*}{\makecell{\textbf{95HD}\textbf{↓}}} & \multirow{3}{*}{\makecell{\textbf{ASD}\textbf{↓}}} \\
\cmidrule{2-4}
& \multirow{2}{*}{\textit{Real}} & \multirow{2}{*}{\textit{Real}} & \multirow{2}{*}{\makecell{\textit{Synthetic}}} & & & & \\
& & & & & & & \\
\midrule
UNet \cite{unet} & 68 & 0 & 0 & 47.83 & 37.01 & 31.16 & 12.62 \\
\midrule
BCP \cite{bai2023bidirectionalcopypastesemisupervisedmedical}$_{\text{\tiny CVPR'23}}$ & \multirow{2}{*}{\makecell{68(5\%)}} & \multirow{2}{*}{\makecell{1244(95\%)}} & \multirow{2}{*}{0} & 87.59 & 78.67 & 1.90 & 0.67 \\
SRA-Seg (Ours)& & & & 87.62 & 78.68 & 2.01 & 0.64 \\
\midrule
BCP \cite{bai2023bidirectionalcopypastesemisupervisedmedical}$_{\text{\tiny CVPR'23}}$ & \multirow{6}{*}{\makecell{68(5\%)}} & \multirow{6}{*}{0} & \multirow{6}{*}{\makecell{1244(95\%)}} & 54.64 & 41.20 & \underline{15.78} & \underline{4.83} \\
CrossMatch \cite{crossmatch}$_{\text{\tiny JBHI'25}}$ & & & & 9.80 & 6.67 & 377.99 & 75.15 \\
ABD(BCP) \cite{abd}$_{\text{\tiny CVPR'24}}$ & & & & 21.84 & 14.83 & \textbf{10.26} & \textbf{1.87} \\
DiffRect \cite{diffrect}$_{\text{\tiny MICCAI'24}}$ & & & & \textbf{63.74} & \textbf{53.51} & 19.29 & 6.13 \\
CGS \cite{cgs}$_{\text{\tiny TMI'25}}$ & & & & 29.23 & 18.93 & 19.20 & 6.23 \\
SRA-Seg (Ours) & & & & \underline{60.02} & \underline{46.52} & 19.95 & 5.83 \\
\midrule\midrule
UNet \cite{unet} & 136 & 0 & 0 & 79.41 & 68.11 & 9.35 & 2.70 \\
\midrule
BCP \cite{bai2023bidirectionalcopypastesemisupervisedmedical}$_{\text{\tiny CVPR'23}}$ & \multirow{2}{*}{\makecell{136(10\%)}} & \multirow{2}{*}{\makecell{1176(90\%)}} & \multirow{2}{*}{0} & 88.84 & 80.62 & 3.98 & 1.17 \\
SRA-Seg (Ours)& & & & 88.99 & 81.00 & 2.82 & 0.83 \\
\midrule
BCP \cite{bai2023bidirectionalcopypastesemisupervisedmedical}$_{\text{\tiny CVPR'23}}$ & \multirow{6}{*}{\makecell{136(10\%)}} & \multirow{6}{*}{0} & \multirow{6}{*}{\makecell{1176(90\%)}} & 87.46 & 78.53 & 5.30 & 1.62 \\
CrossMatch \cite{crossmatch}$_{\text{\tiny JBHI'25}}$ & & & & 85.26 & 76.28 & 3.72 & \underline{1.00} \\
ABD(BCP) \cite{abd}$_{\text{\tiny CVPR'24}}$ & & & & 87.03 & 77.88 & \underline{3.19} & \textbf{0.85} \\
DiffRect \cite{diffrect}$_{\text{\tiny MICCAI'24}}$ & & & & \underline{88.14} & \underline{79.65} & 5.72 & 1.60 \\
CGS \cite{cgs}$_{\text{\tiny TMI'25}}$ & & & & 87.76 & 78.95 & 3.82 & 1.31 \\
SRA-Seg (Ours)& & & & \textbf{89.34} & \textbf{81.24} & \textbf{3.03} & 1.14 \\
\bottomrule
\end{tabular}
\vspace{-5mm}
\end{table}

\begin{figure}
    \centering
    \begin{minipage}{0.102\textwidth}
        \centering
        \includegraphics[width=\linewidth]{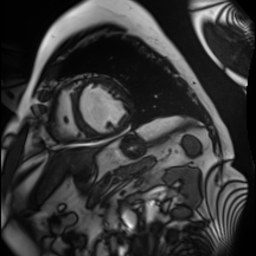}
        \includegraphics[width=\linewidth]{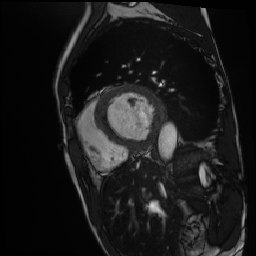}
        \includegraphics[width=\linewidth]{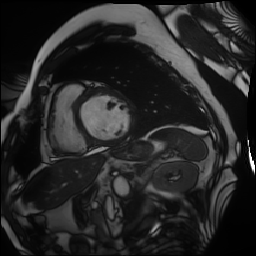}
        \parbox[t][3\baselineskip][t]{\linewidth}{\centering\tiny\textbf{Image}}
    \end{minipage}
    \begin{minipage}{0.102\textwidth}
        \centering
        \includegraphics[width=\linewidth]{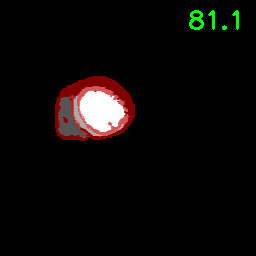}
        \includegraphics[width=\linewidth]{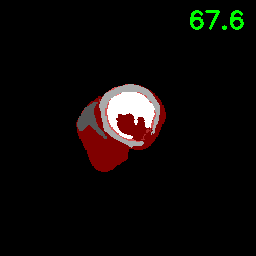}
        \includegraphics[width=\linewidth]{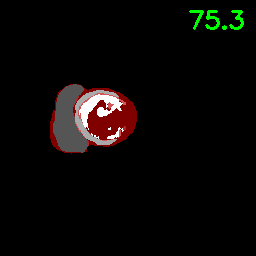}
        \parbox[t][3\baselineskip][t]{\linewidth}{\centering\tiny\textbf{UNet}\tiny\cite{unet}}
    \end{minipage}
    \begin{minipage}{0.102\textwidth}
        \centering
        \includegraphics[width=\linewidth]{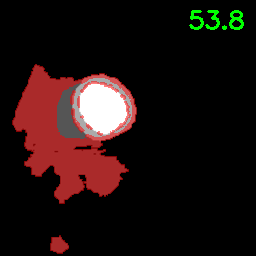}
        \includegraphics[width=\linewidth]{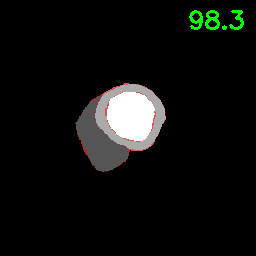}
        \includegraphics[width=\linewidth]{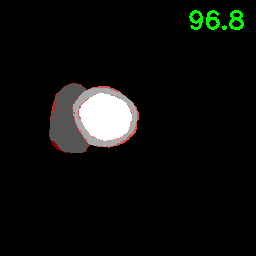}
        \parbox[t][3\baselineskip][t]{\linewidth}{\centering\tiny\textbf{BCP}\tiny\cite{bai2023bidirectionalcopypastesemisupervisedmedical}}
    \end{minipage}
    \begin{minipage}{0.102\textwidth}
        \centering
        \includegraphics[width=\linewidth]{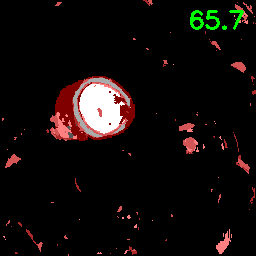}
        \includegraphics[width=\linewidth]{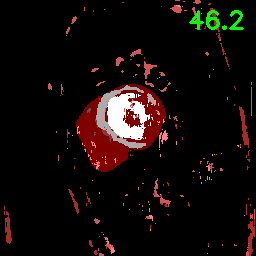}
        \includegraphics[width=\linewidth]{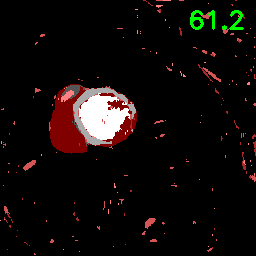}
        \parbox[t][3\baselineskip][t]{\linewidth}{\centering\tiny\textbf{Cross-}\\\tiny\textbf{Match}\tiny\cite{crossmatch}}
    \end{minipage}
    \begin{minipage}{0.102\textwidth}
        \centering
        \includegraphics[width=\linewidth]{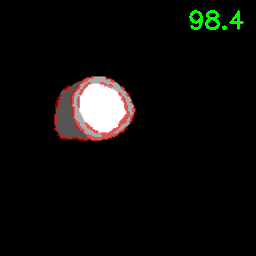}
        \includegraphics[width=\linewidth]{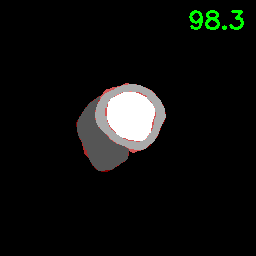}
        \includegraphics[width=\linewidth]{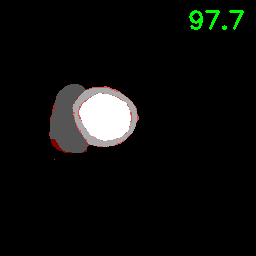}
        \parbox[t][3\baselineskip][t]{\linewidth}{\centering\tiny\textbf{ABD}\\\tiny\textbf{(BCP)}\tiny\cite{abd}}
    \end{minipage}
    \begin{minipage}{0.102\textwidth}
        \centering
        \includegraphics[width=\linewidth]{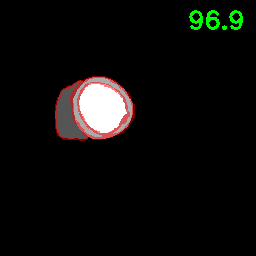}
        \includegraphics[width=\linewidth]{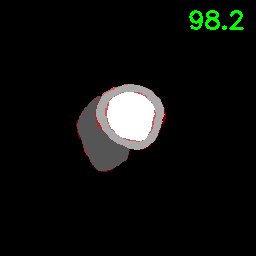}
        \includegraphics[width=\linewidth]{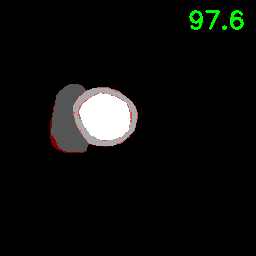}
        \parbox[t][3\baselineskip][t]{\linewidth}{\centering\tiny\textbf{Diff-}\\\tiny\textbf{Rect}\tiny\cite{diffrect}}
    \end{minipage}
    \begin{minipage}{0.102\textwidth}
        \centering
        \includegraphics[width=\linewidth]{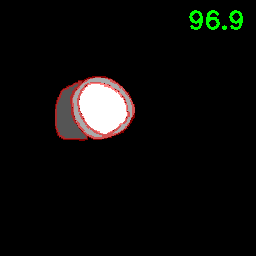}
        \includegraphics[width=\linewidth]{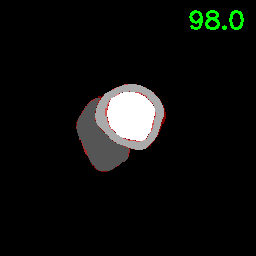}
        \includegraphics[width=\linewidth]{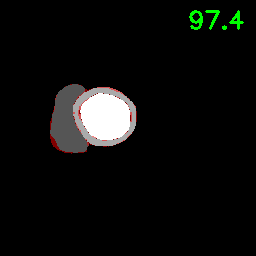}
        \parbox[t][3\baselineskip][t]{\linewidth}{\centering\tiny\textbf{CGS}\tiny\cite{cgs}}
    \end{minipage}
    \begin{minipage}{0.102\textwidth}
        \centering
        \includegraphics[width=\linewidth]{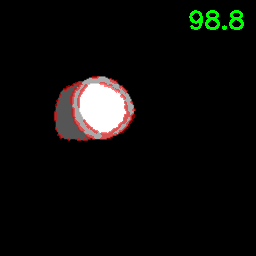}
        \includegraphics[width=\linewidth]{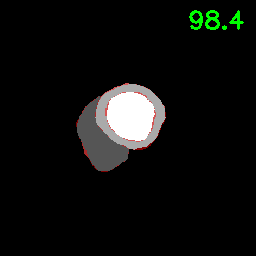}
        \includegraphics[width=\linewidth]{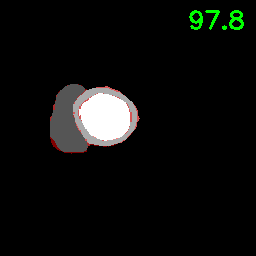}
        \parbox[t][3\baselineskip][t]{\linewidth}{\centering\tiny\textbf{SRA-Seg}}
    \end{minipage}
    \begin{minipage}{0.102\textwidth}
        \centering
        \includegraphics[width=\linewidth]{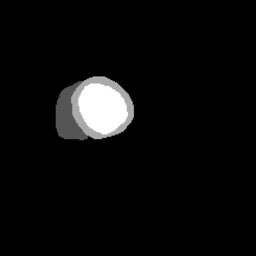}
        \includegraphics[width=\linewidth]{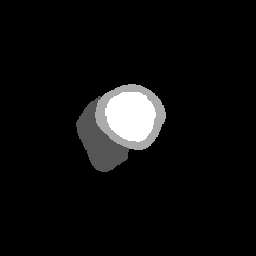}
        \includegraphics[width=\linewidth]{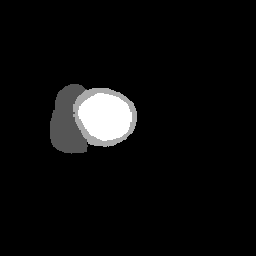}
        \parbox[t][3\baselineskip][t]{\linewidth}{\centering\tiny\textbf{Ground-\\truth}}
    \end{minipage}
    \vspace{-8mm}
    \caption{Qualitative results for 3 images in the ACDC dataset \cite{acdc} considering 10\% real labeled and 90\% synthetic unlabeled data. Each pixel that is incorrectly segmented to a different class is highlighted red. DICE scores are also printed on the top right corner.}
    \label{fig:qual-acdc}
\end{figure}

In the 5\% data split scenario, using synthetic data as opposed to real data dramatically reduces the performance for all approaches. For some methods, the Dice scores fall below 30, rendering these methods essentially unusable, e.g., CrossMatch\cite{crossmatch} drops to 9.80, ABD(BCP)\cite{abd} to 21.84, and CGS\cite{cgs} to 29.23. This poor performance is due to the extremely limited diversity of the 5\% real-labeled subset, which contains data from only three patients with highly similar characteristics. This lack of diversity severely impacts the StyleGAN2-ADA's ability to generate representative synthetic images that capture the full variability of real-world data. Consequently, models trained with this low-diverse synthetic data generalize poorly to the diverse real test set. Section \ref{sec:fid_analysis} provides a detailed analysis using FID scores, which corroborates this observation by quantifying the limited quality and similarity between the synthetic and real data distributions. In contrast, using real unlabeled data leads to superior performance even in this data-constrained 5\% scenario, with our method (SRA-Seg with real unlabeled data) outperforming BCP \cite{bai2023bidirectionalcopypastesemisupervisedmedical} on all metrics except 95HD, where it remains highly competitive (1.90 vs. 2.01).

\noindent \textbf{Qualitative Results:} Figure~\ref{fig:qual-acdc} presents a visual comparison of segmentation results obtained for the 10\% data split on the ACDC dataset. As shown in the overlay highlights, SRA-Seg results in the least errors and achieves significant improvements compared to the other approaches.

\begin{table}[t]
\caption{Comparison of our SRA-Seg approach against other semi-supervised segmentation models on the FIVES dataset \cite{Jin2022}. We mark the best (bold) and second best (underline) results for the synthetic data usage in both 5\% and 10\% data split categories.}
\label{table:quant-fives}
\centering
\scriptsize %
\setlength{\tabcolsep}{1pt} %
\renewcommand{\arraystretch}{0.8}
\begin{tabular}{l|c|cc|c@{\hspace{6pt}}c@{\hspace{6pt}}c@{\hspace{6pt}}c}
\toprule
\multirow{3}{*}{\textbf{Method}} & \textbf{Labeled} & \multicolumn{2}{c|}{\textbf{Unlabeled}} & \multirow{3}{*}{\makecell{\textbf{DICE}\textbf{↑}}} & \multirow{3}{*}{\makecell{\textbf{Jaccard}\textbf{↑}}} & \multirow{3}{*}{\makecell{\textbf{95HD}\textbf{↓}}} & \multirow{3}{*}{\makecell{\textbf{ASD}\textbf{↓}}} \\
\cmidrule{2-4}
& \multirow{2}{*}{\textit{Real}} & \multirow{2}{*}{\textit{Real}} & \multirow{2}{*}{\makecell{\textit{Synthetic}}} & & & & \\
& & & & & & & \\
\midrule
UNet \cite{unet} & 28 & 0 & 0 & 49.50 & 34.38 & 18.03 & 3.55 \\
\midrule
BCP \cite{bai2023bidirectionalcopypastesemisupervisedmedical}$_{\text{\tiny CVPR'23}}$ & \multirow{2}{*}{\makecell{28(5\%)}} & \multirow{2}{*}{\makecell{532(95\%)}} & \multirow{2}{*}{0} & 80.39 & 67.29 & 2.09 & 0.15 \\
SRA-Seg (Ours)& & & & 81.23 & 68.46 & 2.07 & 0.13 \\
\midrule
BCP \cite{bai2023bidirectionalcopypastesemisupervisedmedical}$_{\text{\tiny CVPR'23}}$ & \multirow{6}{*}{\makecell{28(5\%)}} & \multirow{6}{*}{0} & \multirow{6}{*}{\makecell{532(95\%)}} & 82.34 & 70.05 & 1.76 & 0.17 \\
CrossMatch \cite{crossmatch}$_{\text{\tiny JBHI'25}}$ & & & & 62.24 & 45.47 & 3.95 & \textbf{0.03} \\
ABD(BCP) \cite{abd}$_{\text{\tiny CVPR'24}}$ & & & & 70.61 & 54.00 & 3.28 & \underline{0.14} \\
DiffRect \cite{diffrect}$_{\text{\tiny MICCAI'24}}$ & & & & \underline{82.79} & \underline{70.69} & \textbf{1.56} & 0.23 \\
CGS \cite{cgs}$_{\text{\tiny TMI'25}}$ & & & & 81.14 & 68.38 & 1.84 & 0.20 \\
SRA-Seg (Ours)& & & & \textbf{82.85} & \textbf{70.78} & \underline{1.72} & 0.15 \\
\midrule\midrule
UNet \cite{unet} & 56 & 0 & 0 & 59.36 & 43.71 & 13.69 & 2.46 \\
\midrule
BCP \cite{bai2023bidirectionalcopypastesemisupervisedmedical}$_{\text{\tiny CVPR'23}}$ & \multirow{2}{*}{\makecell{56(10\%)}} & \multirow{2}{*}{\makecell{504(90\%)}} & \multirow{2}{*}{0} & 81.87 & 69.35 & 1.85 & 0.15 \\
SRA-Seg (Ours)& & & & 82.58 & 70.38 & 1.80 & 0.16 \\
\midrule
BCP \cite{bai2023bidirectionalcopypastesemisupervisedmedical}$_{\text{\tiny CVPR'23}}$ & \multirow{6}{*}{\makecell{56(10\%)}} & \multirow{6}{*}{0} & \multirow{6}{*}{\makecell{504(90\%)}} & 83.86 & 72.25 & 1.51 & 0.17 \\
CrossMatch \cite{crossmatch}$_{\text{\tiny JBHI'25}}$ & & & & 60.68 & 43.73 & 3.69 & \textbf{0.02} \\
ABD(BCP) \cite{abd}$_{\text{\tiny CVPR'24}}$ & & & & 63.53 & 46.76 & 6.37 & 2.46 \\
DiffRect \cite{diffrect}$_{\text{\tiny MICCAI'24}}$ & & & & \underline{84.22} & \underline{72.79} & \underline{1.41} & \underline{0.16} \\
CGS \cite{cgs}$_{\text{\tiny TMI'25}}$ & & & & 83.55 & 71.80 & 1.59 & 0.17 \\
SRA-Seg (Ours)& & & & \textbf{84.42} & \textbf{73.08} & \textbf{1.34} & \underline{0.16} \\
\bottomrule
\end{tabular}
\vspace{-5mm}
\end{table}

\vspace{-3mm}
\subsection{Results on FIVES Dataset}
\vspace{-2mm}
\textbf{Quantitative Results:} Table~\ref{table:quant-fives} shows that SRA-Seg achieves the best DICE scores among all methods using synthetic unlabeled data in both 5\% and 10\% splits, even outperforming methods using real unlabeled data.

\begin{figure}[t]
    \centering
    \begin{minipage}{0.102\textwidth}
        \centering
        \includegraphics[width=\linewidth]{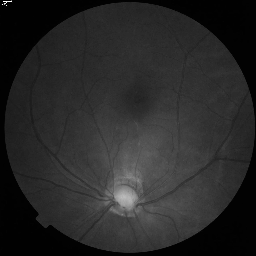}
        \includegraphics[width=\linewidth]{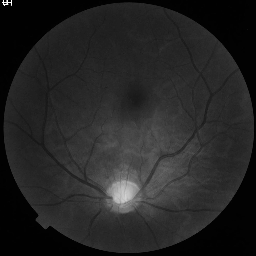}
        \includegraphics[width=\linewidth]{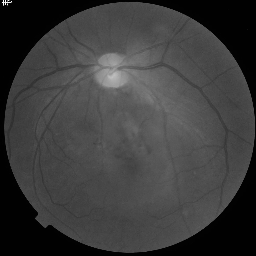}
        \parbox[t][3\baselineskip][t]{\linewidth}{\centering\tiny\textbf{Image}}
    \end{minipage}
    \begin{minipage}{0.102\textwidth}
        \centering
        \includegraphics[width=\linewidth]{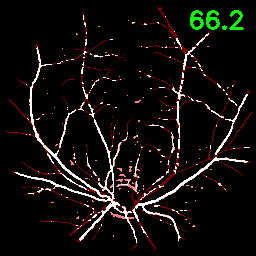}
        \includegraphics[width=\linewidth]{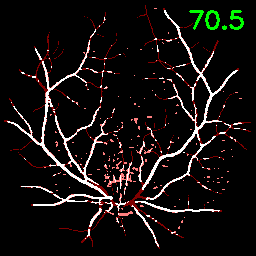}
        \includegraphics[width=\linewidth]{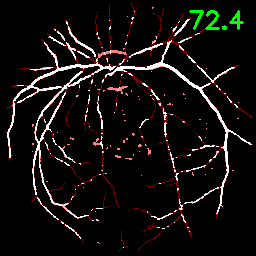}
        \parbox[t][3\baselineskip][t]{\linewidth}{\centering\tiny\textbf{UNet}\tiny\cite{unet}}
    \end{minipage}
    \begin{minipage}{0.102\textwidth}
        \centering
        \includegraphics[width=\linewidth]{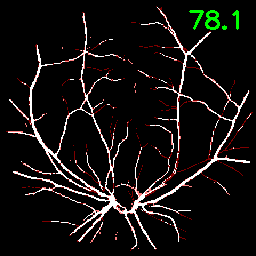}
        \includegraphics[width=\linewidth]{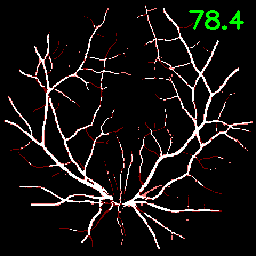}
        \includegraphics[width=\linewidth]{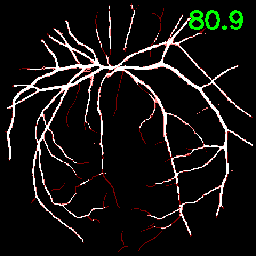}
        \parbox[t][3\baselineskip][t]{\linewidth}{\centering\tiny\textbf{BCP}\tiny\cite{bai2023bidirectionalcopypastesemisupervisedmedical}}
    \end{minipage}
    \begin{minipage}{0.102\textwidth}
        \centering
        \includegraphics[width=\linewidth]{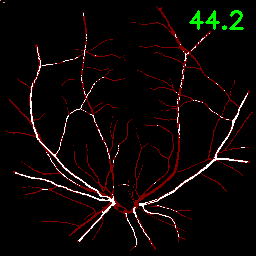}
        \includegraphics[width=\linewidth]{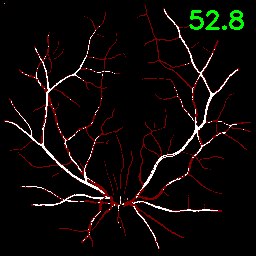}
        \includegraphics[width=\linewidth]{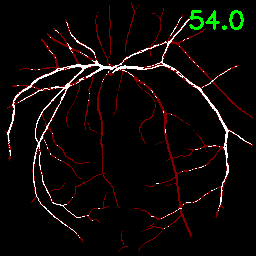}
        \parbox[t][3\baselineskip][t]{\linewidth}{\centering\tiny\textbf{Cross-}\\\tiny\textbf{Match}\tiny\cite{crossmatch}}
    \end{minipage}
    \begin{minipage}{0.102\textwidth}
        \centering
        \includegraphics[width=\linewidth]{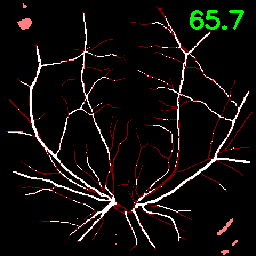}
        \includegraphics[width=\linewidth]{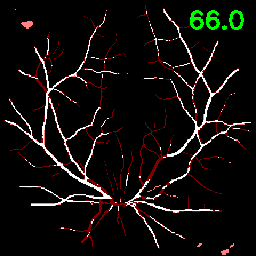}
        \includegraphics[width=\linewidth]{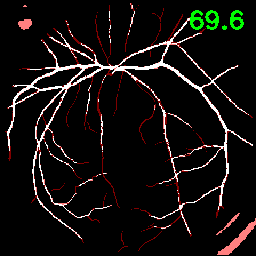}
        \parbox[t][3\baselineskip][t]{\linewidth}{\centering\tiny\textbf{ABD}\\\tiny\textbf{(BCP)}\tiny\cite{abd}}
    \end{minipage}
    \begin{minipage}{0.102\textwidth}
        \centering
        \includegraphics[width=\linewidth]{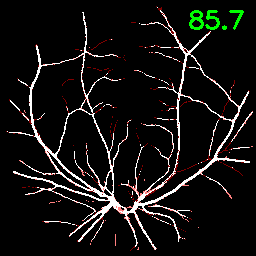}
        \includegraphics[width=\linewidth]{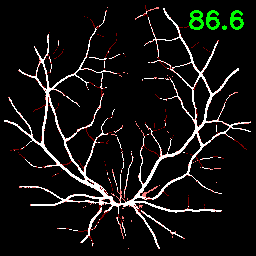}
        \includegraphics[width=\linewidth]{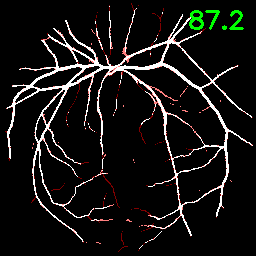}
        \parbox[t][3\baselineskip][t]{\linewidth}{\centering\tiny\textbf{Diff-}\\\tiny\textbf{Rect}\tiny\cite{diffrect}}
    \end{minipage}
    \begin{minipage}{0.102\textwidth}
        \centering
        \includegraphics[width=\linewidth]{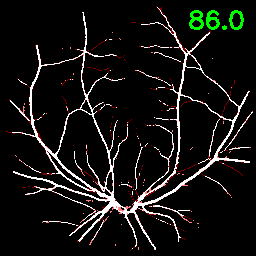}
        \includegraphics[width=\linewidth]{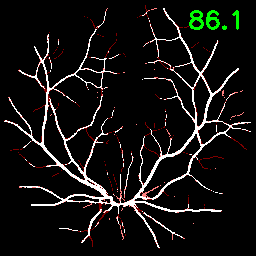}
        \includegraphics[width=\linewidth]{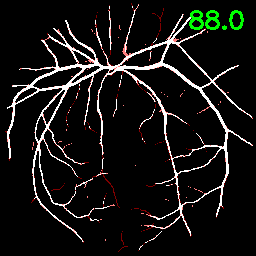}
        \parbox[t][3\baselineskip][t]{\linewidth}{\centering\tiny\textbf{CGS}\tiny\cite{cgs}}
    \end{minipage}
    \begin{minipage}{0.102\textwidth}
        \centering
        \includegraphics[width=\linewidth]{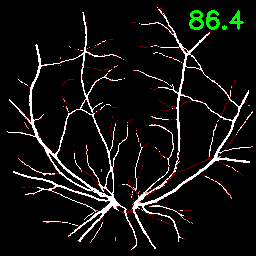}
        \includegraphics[width=\linewidth]{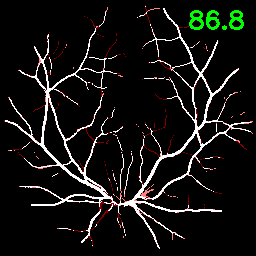}
        \includegraphics[width=\linewidth]{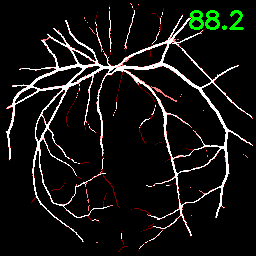}
        \parbox[t][3\baselineskip][t]{\linewidth}{\centering\tiny\textbf{SRA-Seg}}
    \end{minipage}
    \begin{minipage}{0.102\textwidth}
        \centering
        \includegraphics[width=\linewidth]{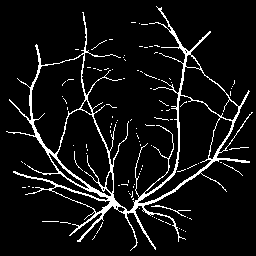}
        \includegraphics[width=\linewidth]{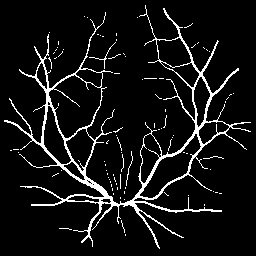}
        \includegraphics[width=\linewidth]{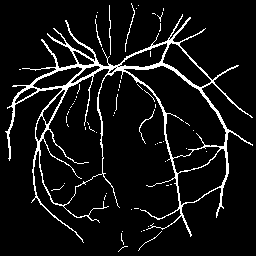}
        \parbox[t][3\baselineskip][t]{\linewidth}{\centering\tiny\textbf{Ground-\\truth}}
    \end{minipage}
    \vspace{-8mm}
    \caption{Qualitative results for 3 images in the FIVES dataset \cite{Jin2022} considering 10\% real labeled and 90\% synthetic unlabeled data. Each pixel that is incorrectly segmented to a different class is highlighted red. DICE scores are also printed on the top right corner.}
    \label{fig:qual-fives}
    \vspace{-6mm}
\end{figure}

\noindent \textbf{Qualitative Results:} Figure~\ref{fig:qual-fives} provides qualitative segmentation results on the FIVES dataset. Consistent with our quantitative analysis, these visualizations highlight SRA-Seg's superior performance in accurately segmenting various anatomical structures compared to other approaches.

\vspace{-6mm}
\subsection{Analysis of Synthetic Data Quality (FID Scores)}
\label{sec:fid_analysis}
\vspace{-3mm}
To quantitatively assess fidelity and perceptual similarity between our generated synthetic images and the real counterparts, we utilize Fréchet Inception Distance (FID)~\cite{Seitzer2020FID}. A lower FID score indicates that the distribution of generated images is closer to the real image distribution, signifying higher fidelity and diversity in the synthetic set~\cite{Woodland_2022,heusel2018ganstrainedtimescaleupdate}.

\begin{figure}[b]  %
    \centering
    \vspace{-4mm}
    \includegraphics[width=\linewidth]{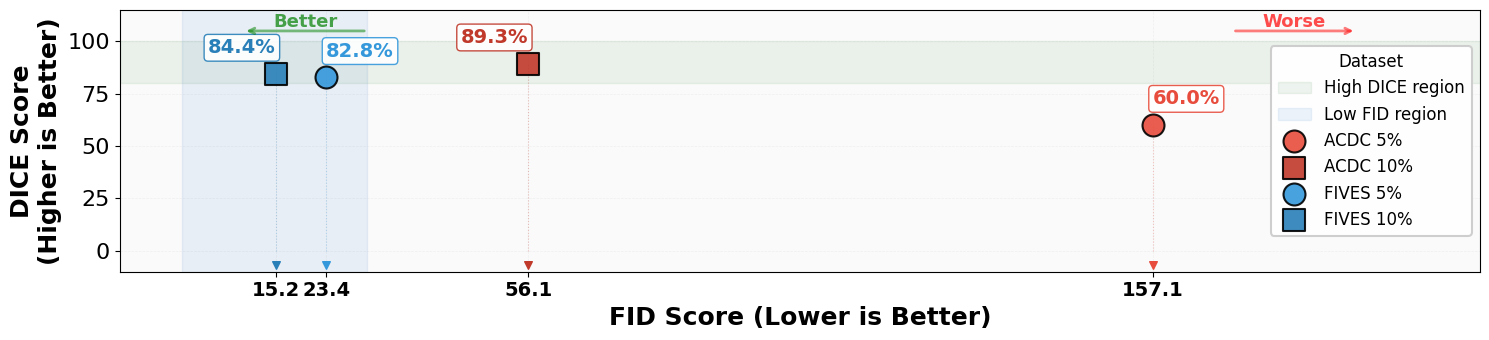}  %
    \vspace{-6mm}
    \caption{FID Scores comparing synthetic and real images across 5\% and 10\% data split in both ACDC\cite{acdc} and FIVES\cite{Jin2022} datasets.}
    \label{fig:fid}  %
\end{figure}

Figure \ref{fig:fid} illustrates FID scores for both datasets across 5\% and 10\% splits. For ACDC, the 5\% split yields a high FID score (157.06), indicating a significant domain gap and lower synthetic image quality compared to real data, while the 10\% split shows improved alignment (56.06), confirming the performance drop observed at 5\%. In contrast, FIVES achieves consistently lower FID scores (23.39 and 15.24 for 5\% and 10\%, respectively), suggesting its synthetic data is inherently more realistic and better aligned with real distributions.

Overall, these FID results underscore that synthetic data quality is paramount for effective semi-supervised learning. While increasing the labeled fraction generally improves synthetic data fidelity and domain alignment, the baseline diversity of the dataset significantly influence the generative model's ability to produce usable synthetic images. Using FID score serves as a crucial indicator of the quality and suitability of synthetic data for methods like SRA-Seg.

\subsection{Ablation Studies}
\vspace{-2mm}
To assess the contribution of each component, we conducted ablation experiments on both ACDC~\cite{acdc} and FIVES~\cite{Jin2022} datasets using 10\% real labeled and 90\% synthetic unlabeled data. Table~\ref{table:ablation} report segmentation performance under different configurations of our three key components: soft-mix augmentation, soft-loss, and SA-loss. These studies clearly demonstrate that each proposed element provides consistent gains across DICE, Jaccard, 95HD, and ASD metrics, with the combination of all three components achieving the best performance.

\begin{table}[t]
\caption{Ablation study on ACDC \cite{acdc} and FIVES \cite{Jin2022}.}
\label{table:ablation}
\centering
\scriptsize
\setlength{\tabcolsep}{1pt}
\renewcommand{\arraystretch}{0.8}
\begin{tabular}{l|c@{\hspace{4pt}}c@{\hspace{4pt}}c|c@{\hspace{4pt}}c@{\hspace{4pt}}c@{\hspace{4pt}}c|c@{\hspace{4pt}}c@{\hspace{4pt}}c@{\hspace{4pt}}c}
\toprule
& & & & \multicolumn{4}{c|}{\textbf{ACDC}} & \multicolumn{4}{c}{\textbf{FIVES}} \\
\cmidrule(lr){5-8} \cmidrule(lr){9-12}
\textbf{Method} & \makecell{\textbf{Soft-}\\\textbf{mix}} & \makecell{\textbf{Soft-}\\\textbf{loss}} & \makecell{\textbf{SA-}\\\textbf{loss}} & \makecell{\textbf{DICE↑}} & \makecell{\textbf{Jaccard↑}} & \makecell{\textbf{95HD↓}} & \makecell{\textbf{ASD↓}} & \makecell{\textbf{DICE↑}} & \makecell{\textbf{Jaccard↑}} & \makecell{\textbf{95HD↓}} & \makecell{\textbf{ASD↓}} \\
\midrule
BCP     & \ding{53} & \ding{53} & \ding{53} & 87.46 & 78.53 & 5.30 & 1.62 & 83.86 & 72.25 & 1.51 & 0.17 \\
        & \ding{53} & \ding{53} & \ding{51} & 87.60 & 78.66 & 9.66 & 2.75 & 84.02 & 72.50 & 1.51 & 0.16 \\
        & \ding{51} & \ding{53} & \ding{53} & 88.16 & 79.42 & 4.80 & 1.50 & 84.08 & 72.58 & 1.54 & 0.17 \\
        & \ding{51} & \ding{53} & \ding{51} & 88.09 & 79.36 & \textbf{2.14} & \textbf{0.78} & 84.25 & 72.83 & 1.44 & 0.16 \\
        & \ding{51} & \ding{51} & \ding{53} & 88.96 & 80.69 & 2.87 & 1.01 & 84.16 & 72.71 & 1.51 & 0.16 \\
SRA-Seg & \ding{51} & \ding{51} & \ding{51} & \textbf{89.33} & \textbf{81.25} & 2.94 & 0.85 & \textbf{84.42} & \textbf{73.08} & \textbf{1.34} & \textbf{0.16} \\
\bottomrule
\end{tabular}
\vspace{-6mm}
\end{table}

\vspace{-5mm}
\section{Conclusion}
\vspace{-3mm}
SRA-Seg demonstrates that synthetic data, when properly aligned with real distributions, can serve as an effective substitute for real unlabeled data in semi-supervised medical image segmentation. Our method effectively addresses the synthetic-real domain gap through principled feature alignment and augmentation strategies. Experimental results validate our approach: using only 10\% real labeled data with 90\% synthetic data, SRA-Seg improves Dice scores by 1.88 on ACDC\cite{acdc} and 0.56 on FIVES\cite{Jin2022} over BCP\cite{bai2023bidirectionalcopypastesemisupervisedmedical}, outperforming state-of-the-art semi-supervised methods and matching the performance of approaches trained entirely on real data. While extremely low-diversity scenarios remain challenging due to limitations in current generative models, this work establishes a principled foundation for leveraging synthetic data to overcome the annotation bottleneck that has long constrained medical imaging AI. As generative models continue to advance, SRA-Seg provides a ready framework to harness their improvements for more accurate and accessible medical image segmentation.
\bibliographystyle{splncs04}
\bibliography{references}
\end{document}